\documentclass[letterpaper, 10 pt, journal, twoside]{IEEEtran}
\usepackage{times}

\usepackage[numbers]{natbib}
\usepackage{multicol,balance}
\usepackage[bookmarks=true]{hyperref}
\usepackage{amsmath} 
\usepackage[italic]{mathastext}

\DeclareMathOperator*{\argmin}{arg\,min}
\usepackage{bbm}
\usepackage{tcolorbox}
\DeclareMathOperator{\conf}{con\hspace{-0.1em}f}
\DeclareMathOperator{\bconf}{\textbf{con}\hspace{-0.1em}\textbf{f}}

\BeforeBeginEnvironment{equation}{\par\vspace{-.8em}} 
\AfterEndEnvironment{equation}{\vspace{-.8em}\par} 

\usepackage{amssymb}

\usepackage{xcolor} 

\usepackage{algorithm}
\usepackage[noEnd]{algpseudocodex} 

\let\labelindent\relax
\usepackage{enumitem}

\usepackage{booktabs}
\usepackage{siunitx}

\IEEEoverridecommandlockouts

\usepackage{fancyhdr}
\fancypagestyle{plain}{
  \fancyhf{}
  \fancyhead[C]{Conference on \LaTeX}     
  \fancyfoot[L]{This is a notice}

}
\usepackage{eso-pic}
\usepackage{hyperref}

\begin{document}

\AddToShipoutPictureBG*{%
  \AtPageLowerLeft{%
    \setlength\unitlength{1in}%
    \hspace*{\dimexpr0.5\paperwidth\relax}
    \makebox(0,0.75)[c]{\parbox{0.7\paperwidth}{\raggedright\footnotesize© 2023 IEEE.  Personal use of this material is permitted. Permission from IEEE must be obtained for all other uses, in any current or future media, including reprinting/republishing this material for advertising or promotional purposes, creating new collective works, for resale or redistribution to servers or lists, or reuse of any copyrighted component of this work in other works.}}%
}}

\title{Coordinated Multi-Robot Shared Autonomy Based on Scheduling and Demonstrations}

\markboth{IEEE Robotics and Automation Letters. Preprint Version. Accepted October, 2023}
{Hagenow \MakeLowercase{\textit{et al.}}: Coordinated Multi-Robot Shared Autonomy Based on Scheduling and Demonstrations} 

\author{Michael Hagenow,$^{1}$ Emmanuel Senft,$^{2}$ Nitzan Orr,$^{3}$ Robert Radwin,$^{4}$ \\ Michael Gleicher,$^{3}$ Bilge Mutlu,$^{3}$ Dylan P. Losey,$^{5}$ Michael Zinn$^{1}$
\thanks{Manuscript received: March, 25, 2023; Revised July, 31, 2023; Accepted October, 11, 2023.}
\thanks{This paper was recommended for publication by Editor M. Ani Hsieh upon evaluation of the Associate Editor and Reviewers' comments.}
\thanks{This work was supported in part by the Grainger Wisconsin Distinguished Graduate Fellowship (WDGF) and in part by a NASA University Leadership Initiative (ULI) grant awarded to the UW-Madison and The Boeing Company (Cooperative Agreement \# 80NSSC19M0124).}
\thanks{$^{1}$Michael Hagenow and Michael Zinn are with the Department of Mechanical
Engineering, University of Wisconsin--Madison, Madison 53706, USA
        {\tt\small [mhagenow|mzinn]@wisc.edu}}%
\thanks{$^{2}$Emmanuel Senft is a Research Scientist at the Idiap Research Institute, Martigny, Switzerland
        {\tt\small esenft@idiap.ch}}
\thanks{$^{3}$Nitzan Orr, Michael Gleicher, and Bilge Mutlu are with the Department of Computer
Sciences, University of Wisconsin--Madison, Madison 53706, USA
        {\tt\small [nitzan|gleicher|bilge]@cs.wisc.edu}}
\thanks{$^{4}$Robert Radwin is with the Department of Industrial and Systems Engineering, University of Wisconsin--Madison, Madison 53706, USA
        {\tt\small rradwin@wisc.edu}}
\thanks{$^{5}$Dylan P. Losey is with the Department of Mechanical Engineering, Virginia Tech, Blacksburg 24061, USA
        {\tt\small losey@vt.edu}}
\thanks{Digital Object Identifier (DOI): \href{https://doi.org/10.1109/LRA.2023.3327625}{10.1109/LRA.2023.3327625}.}
}


%

\maketitle

\begin{abstract}
Shared autonomy methods, where a human operator and a robot arm work together, have enabled robots to complete a range of complex and highly variable tasks. Existing work primarily focuses on one human sharing autonomy with a \textit{single} robot. By contrast, in this paper we present an approach for \textit{multi-robot shared autonomy} that enables one operator to provide real-time corrections across two coordinated robots completing the same task in parallel. Sharing autonomy with multiple robots presents fundamental challenges. The human can only correct one robot at a time, and without coordination, the human may be left idle for long periods of time. Accordingly, we develop an approach that aligns the robot's learned motions to best utilize the human's expertise.  Our key idea is to leverage \textit{Learning from Demonstration (LfD)} and \textit{time warping} to schedule the motions of the robots based on when they may require assistance. Our method uses variability in operator demonstrations to identify the types of corrections an operator might apply during shared autonomy, leverages flexibility in how quickly the task was performed in demonstrations to aid in scheduling, and iteratively estimates the likelihood of when corrections may be needed to ensure that only one robot is completing an action requiring assistance. Through a preliminary study, we show that our method can decrease the scheduled time spent sanding by iteratively estimating the times when each robot could need assistance and generating an optimized schedule that allows the operator to provide corrections to each robot during these times.
\end{abstract}


\IEEEpeerreviewmaketitle


\section{Introduction}
\label{sec:introduction}

\begin{figure}[t]
\centering
\includegraphics[width=3.4in]{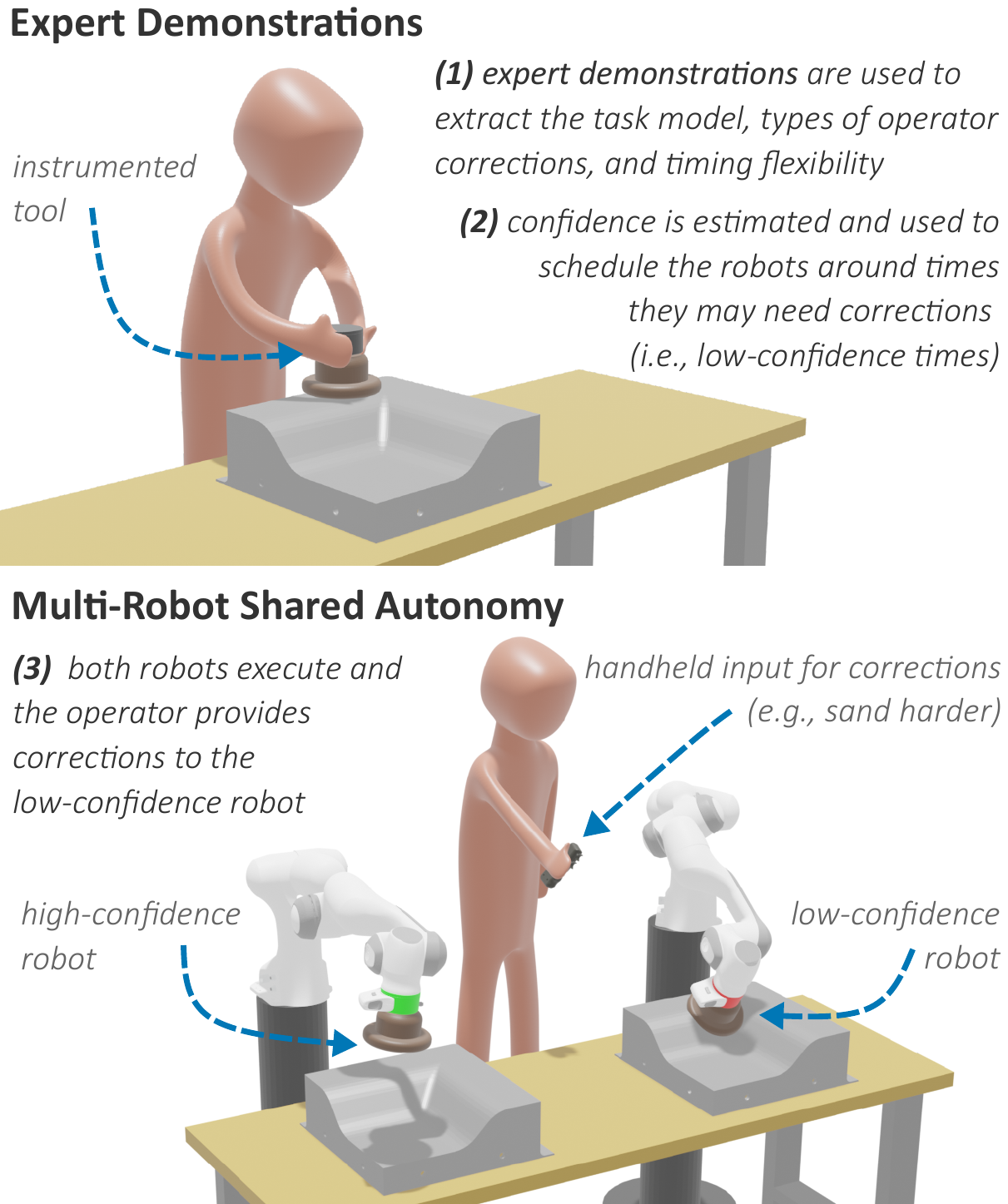}
\vspace{-5pt}
\caption{Our method enables \textit{multi-robot shared autonomy} by leveraging expert demonstrations and coordinating robots around times where they may need assistance from the operator. \textit{Top:} An operator (task expert) provides a set of demonstrations to inform the robot task model, task variability, and flexibility in scheduling. \textit{Bottom:} The operator provides corrections across two robots completing the same task. The robots are scheduled such that only one robot (\textit{e.g.}, the robot with the red indicator) could require corrections at any given time (\textit{i.e.}, the other robot has high confidence in its action). The confidence is estimated from demonstration variability and empirical correction data.}
\label{fig:teaser}
\vspace{-20pt}
\end{figure}

\IEEEPARstart{H}{uman}-robot teaming is a promising alternative for tasks where robust automation is infeasible due to high complexity and variability.
In many cases, human-robot teaming is achieved through \textit{shared autonomy}, where a human operator and a robot policy share command over the physical robot platform and leverage their respective strengths \cite{selvaggio2021autonomy,losey2018review}. For example, the expert operator can offload a task's physical burden to the robot and use their own task knowledge and superior sensing to make corrections to the robot's behavior. However, many tasks do not require input from the operator during the entire execution but only during \emph{regions of task variability}. For example, during a fastener insertion task, the robot may only need help from the human to fix alignment error when inserting fasteners but not while fetching or prepping the fasteners. When the regions of variability make up a small part of the overall task the worker is poorly utilized. While the worker may perform secondary tasks during idle time, there are advantages to the operator working with multiple robots engaging in the same task, such as reducing context switching. In this work, we propose a method for multi-robot shared autonomy that sequences the execution of two robots based on regions where they may require assistance.

Previous work includes investigations of elements of multi-robot shared autonomy, such as interfaces for operator attention management \cite{gao2012teamwork}, supervisor allocation across a fleet of agents (e.g., mobile robots) \cite{crandall2010computing,rosenfeld2017intelligent,dahiya2022scalable,swamy2020scaled,hoque2022fleet,ji2022traversing}, and scheduling of agent subtasks and supervision \cite{yan2013survey,gombolay2013fast,cai2022scheduling,zanlongo2021scheduling}. However, the allocation methods focus on enabling an operator to temporarily teleoperate an agent needing assistance and in scheduling, little to no work focuses on coordinating agents around operator intervention. As illustrated in Figure \ref{fig:teaser}, we are interested in \textit{coordinated shared autonomy} where the robots operate at a high level of autonomy. In this paradigm, the robots sequence their behaviors such that only one robot could require assistance at any point, and the operator provides targeted corrections to the low-confidence robot if its action is incorrect. The major challenges in coordinated shared autonomy are determining \emph{what} interventions the operator may want to provide and \emph{when} they may occur during the task. Addressing these challenges requires models of the task, the types of corrections the operator may desire, and when corrections might be needed. Scheduling coordinated shared autonomy also introduces new challenges addressed in this work, such as how to compensate for changes to the timing of the robot's execution that result from operator corrections.

In this paper, we propose a method for multi-robot shared autonomy based on operator corrections \cite{hagenow2021corrective}, multi-agent scheduling, and Learning from Demonstration (LfD) \cite{ravichandar2020recent}. Our method schedules two robots such that one operator can provide real-time corrections at times when they are needed by both robots. The task model, corrections an operator can make, and sequencing of agents are inferred from expert demonstrations and past shared autonomy executions. The main contributions of this paper include (1) an optimization-based method to schedule corrections-based shared autonomy of multiple agents by leveraging variability in demonstrations; (2) a technique for inferring when corrections are needed based on task variability and Bayesian inference; and (3) a real-time adaptation strategy to update the timing of robots when operator corrections cause deviations from the schedule.

\section{Optimizing for Multi-Robot Corrections}
\label{sec:method}

Our method is formulated as an optimization problem and enables one human operator to provide real-time corrections to two robots completing different instances of the same task.

\begin{figure*}[t]
\centering
\includegraphics[width=\textwidth]{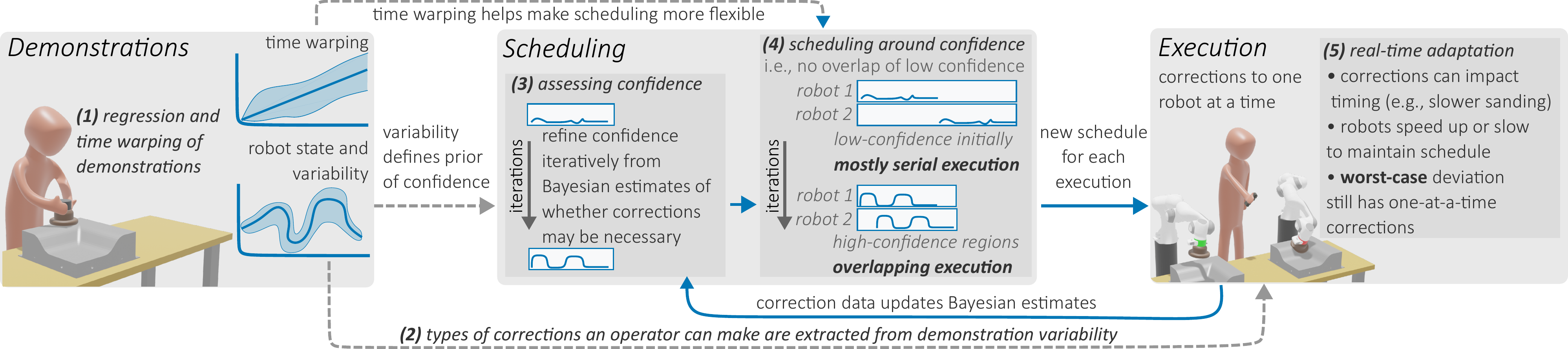}
\vspace{-15pt}
\caption{High-level overview of the proposed method. The numbers correspond to the subsection of \S\ref{sec:approach} where the technical details are described. Blue arrows indicate the flow of information at every iteration. Gray arrows indicate information that is only processed once.}
\label{fig:methodoverview}
 \vspace{-15pt}
\end{figure*}

\subsection{Problem Setting and Assumptions}
We explore settings where there are variable aspects of the task that the robots \textit{cannot learn} due to sensor limitations. For example, an expert sander may at times use subtle cues (e.g., watching material build up on the side of the tool, changing views of the surface) to choose an appropriate action. 
We consider settings with two robots and one skilled operator completing a high-volume (i.e., repetitive) task where the types of variability encountered do not change over time. There are many industrial tasks that meet this requirement. For example, in many sanding applications, there are small regions of the workpiece that will have defects from upstream manufacturing processes which will cause each piece to have different sanding needs.
To encode the robot behaviors, we assume access to a set, $\mathcal{D}$, of $n_d$ varying-length expert demonstrations.
\begin{equation}
    \begin{aligned}
        \mathcal{D} = \{ \textbf{X}_1, ... \textbf{X}_{n_d} \} \\
        \forall \textbf{X}_{i} \in \mathcal{D},\, \textbf{X}_i \in \mathbb{R}^{m \times T_i} 
    \end{aligned}
\end{equation}
where $\textbf{X}_i$ is the demonstration state data (e.g., motions, forces, tool state) with $m$ state variables and $T_{i}$ is the length of the demonstration. For brevity, we will use subscripts on the demonstration set to refer to the data of an individual demonstration (i.e., $\mathcal{D}_i \equiv \textbf{X}_i$) and refer to interpolated values of the demonstration data using function notation (i.e., $\textbf{x}_i(t)$). To enable time warping and regression across multiple demonstrations, we assume the demonstrations are performed using similar trajectories (e.g., always going left to right when sanding). We also require that the robots must execute the task as shown in the demonstrations to satisfy any sequential constraints of the task. In other words, the execution cannot be broken up or reordered. Finally, we assume the robots must operate within the demonstration execution rates (i.e., velocities). While in some tasks it may be possible for the robots to stop at times when they need help, there are many examples where this is not possible. For example, a robot that needs help with part of a sanding pass cannot stop mid-pass without impacting the continuity of the sanding.

\subsection{Proposed Approach}\label{sec:approach}
Our approach centralizes around scheduling the two robot executions. Our high-level method is illustrated in Figure \ref{fig:methodoverview}. We start with a set of expert demonstrations to expose the task variability, including differences in how quickly the task was executed (i.e., the execution rate), and encode the robot behavior. To schedule the robots around operator corrections, we need to know what types of corrections the operator wishes to provide and when they may be needed. Following previous work \cite{hagenow2021informing}, we assume that the demonstration variance indicates the corrections that may be necessary (e.g., variance in force indicates potential force corrections) and allow operators to provide these differential corrections by mapping the variance to a low degree-of-freedom input (e.g., joystick).

We determine when the corrections may be needed by estimating the robot confidence throughout the task. Specifically, we identify times in the robot behavior that might require corrections (i.e., low-confidence times) and times where corrections are unlikely (i.e., high-confidence times). The two robot behaviors are scheduled such that the low-confidence actions do not occur at the same time. The robot confidence is estimated iteratively. As the operator uses the system, it collects empirical data of when the operator provides corrections and refines its confidence estimates. Each time the system is used, we update the confidence and re-optimize the schedule based on the new confidence estimates. As the amount of high-confidence times increases, the scheduling method finds more efficient schedules by shortening and overlapping the robot executions. Our scheduling method leverages flexibility in the execution rate during high-confidence times to aid in finding an optimized schedule. As operator corrections can impact the timing of the robots, we also ensure the scheduling solution is robust to timing deviations from the corrections. The following subsections describe our method's technical details.

\subsubsection{Regression and Time Warping of Demonstrations}
To encode the desired robot skill and to uncover time flexibility in the demonstrations, we take advantage of \textit{time warping} \cite{zhou2012generalized}.
A time warp, $\psi$, is a nonlinear and monotonically increasing function that maps a time between two sets of data:
\begin{equation}
    \psi_{\mathcal{D}_1 \rightarrow \mathcal{D}_2}: \left[ 0, T_1 \right] \rightarrow \left[ 0 ,T_2 \right]
\end{equation}
\vspace{-.3cm}
This is accomplished by minimizing a loss between the data of the two time series. We consider a weighted metric that accounts for state variables with varying units. The time warping optimization can be summarized as:
\begin{equation}
\label{eq:warp}
\begin{aligned}
    \min_{\psi_{\mathcal{D}_1 \rightarrow \mathcal{D}_2}} \int^{T_1}_{0}\mathcal{L}(\textbf{x}_{2}(\psi_{\mathcal{D}_1 \rightarrow \mathcal{D}_2}(t))-\textbf{x}_{1}(t)) dt \\
    \mathcal{L}(\textbf{e}_{\psi}) = \textbf{e}_{\psi}^{\intercal}\textbf{W}\textbf{e}_{\psi}
\end{aligned}
\end{equation}
 where $\textbf{e}_{\psi}$ is the error between the vectors of the two states and $\textbf{W} \in \mathbb{R}^{m \times m}$ is a diagonal weighting matrix with nonnegative weights that is set based on variance or domain knowledge. We solve Equation \ref{eq:warp} using a parametric optimizer \cite{deriso2022general}. We align all demonstrations to the first demonstration as a common reference and thus truncate the warp notation going forward by dropping the reference demonstration (e.g., $\psi_{\mathcal{D}_2} \equiv \psi_{\mathcal{D}_1 \rightarrow \mathcal{D}_2}$). Given that we will optimize new warps for each robot execution, the choice of reference is mostly arbitrary. Going forward, we mainly focus on the gradients of the warps. Working with the gradient allows us to assess the rate at which the robot can execute the task at any given time and allows us to easily enforce that warps are monotonically increasing. 

Using the time warps, we align the demonstration data to determine the robot behavior. There are many possible methods for cloning the demonstration data. In this work, we use a simple regression that estimates of the mean behavior of the aligned demonstrations, which we denote as $\hat{\boldsymbol{\mu}}(t)$.

\subsubsection{Process Variability and Operator Corrections}
During a robot's execution, the operator can provide real-time corrections to a robot's behavior. We provide a simplified control to the operator by extracting the likely types of corrections from the variability in the expert demonstrations. We leverage our previous method \cite{hagenow2021informing} to determine admissible corrections, including coordinations of robot state variables (e.g., pitch, force, and speed to modulate abrasiveness). The remainder of this subsection provides a brief review of the previous method for completeness.

The differential corrections are provided through an input device that functions like a joystick. Given that the corrections can affect the execution rate, we consider an augmented robot state that includes the warp gradient.
\begin{equation}
 \textbf{x}_i^{\textrm{+}}(t) = [ \textbf{x}_i^{\intercal}(t), \dot{\psi}_{\mathcal{D}_i}(t) ]^{\intercal}
\end{equation}
\vspace{-.3cm}
where the superscript $+$ refers to the augmented state. The augmented mean behavior is also defined by concatenating the mean warp gradient of the demonstrations.
\begin{equation}
\begin{aligned}
\dot{\bar{\psi}}_{\mathcal{D}}(t) = \sum\limits_{i}^{n_d} \dot{\psi}_{\mathcal{D}_i}(t) / n_d \\
\hat{\boldsymbol{\mu}}^{\textrm{+}}(t) = [ \hat{\boldsymbol{\mu}}^{\intercal}(t), \dot{\bar{\psi}}_{\mathcal{D}}(t) ]^{\intercal}
\end{aligned}
\end{equation}
where $\dot{\bar{\psi}}_{\mathcal{D}}(t)$ is the mean warp gradient and $\hat{\boldsymbol{\mu}}^{+}(t)$ is the augmented mean behavior. The mean-removed data and demonstration variability can subsequently be calculated:
\begin{equation}
    \label{eq:variancefromdemos}
    \begin{aligned}
     \textbf{e}_i^{\textrm{+}}(t) = \textbf{x}_{i}^{\textrm{+}}\left(\psi_{\mathcal{D}_i}(t)\right)-\hat{\boldsymbol{\mu}}^{+}(t) \\
    \sigma^2_{\mathcal{D}}(t) = \frac{1}{n_d-1}\sum \limits_{i} \left(\textbf{e}_i^{\textrm{+}}(t) \right) ^{\intercal}\textbf{W}^{+}\left( \textbf{e}_i^{\textrm{+}}(t) \right)
    \end{aligned}
\end{equation}
where $\textbf{e}_i^{\textrm{+}}(t)$ is the mean-removed data, $\sigma^2_{\mathcal{D}}(t)$ is the magnitude of the demonstration variance, and $\textbf{W}^{\textrm{+}} \in \mathbb{R}^{(m+1) \times (m+1)}$ is the expanded diagonal weighting matrix that adds a weight for the warp gradient variable. The final operator correction, $\mathbf{\delta} \textbf{x}^{+}(u,t)$, is based on the principal components at a given time (using principal component analysis) and the user input, $u$. The maximum magnitude of the correction is scaled such that the corrections do not exceed the permissible warp gradients from the demonstrations. The final robot command is arbitrated as the sum of the mean state and differential operator correction.
\begin{equation}
        \textbf{x}_{f}^{+}(u,t) = \hat{\boldsymbol{\mu}}^{+}(t) + \delta \textbf{x}^{+}(u,t)
\label{eq:arbitration}
\end{equation}
\vspace{-.3cm}
where $\textbf{x}^{+}_{f}$ is the final robot state. The correction also determines the next robot time based on the warp gradient. The timing of the robots is discussed in detail in a later section.

\subsubsection{Assessing Robot Confidence}
Intuitively, we want the robots to overlap at times when at least one robot is confident and \textit{not} to overlap when they are both unsure. This scheduling requires determining the times of the execution where the robot may require corrections (i.e., the robot has low confidence in its action). Let $p_{c\mid t}$ be the probability of the expert giving a correction to a robot at a given time. We desire to estimate $p_{c\mid t}$ and then threshold the value to determine whether a robot is low or high confidence. While variability in the demonstrations serves as an indicator of what corrections an operator may provide, this estimate is too conservative for estimating $p_{c\mid t}$ given the multiple sources of demonstration variability \cite{todorov2002minimal}. In some cases, the variability is from task uncertainty (e.g., different forces depending on paint when sanding). However, there are also cases when demonstration variability is non-critical. For example, an operator's tool path when moving between sanding passes is non-critical and underconstrained, and thus may be variable.

Our solution leverages the demonstration variability and empirical correction data from past executions of the system to iteratively estimate the probability of corrections at every time step. Initially, the system will only be high-confidence for times where there was negligible variability in the human demonstrations. However, as the system collects empirical data from the $n_e$ previous executions of whether corrections are provided, the system will infer additional high-confidence times and can schedule shorter, more parallel solutions to reduce the overall time on task. The empirical data, $\textbf{z}_t \in \mathbb{B}^{n_e}$, is a set for each time step (defined in the reference demonstration) of whether corrections were given during each robot execution. With this empirical data, we use Bayesian inference to estimate the probability distribution that a correction is needed.
\begin{equation}
    P(p_{c\mid t} \mid \textbf{z}_t) = \frac{P(\textbf{z}_t \mid p_{c\mid t})P(p_{c\mid t})}{P(\textbf{z}_t)}
\end{equation}
Given the empirical data is Boolean and assuming the events (i.e., executions) are independent, this inference problem can be modeled using a Beta distribution \cite{gupta2004handbook}. The Bayesian posterior can be inferred based on a prior and the empirical data:
\begin{equation}
    P(p_{c\mid t} \mid \textbf{z}_t) = beta(\alpha_{0,t} + \sum \textbf{z}_t, \hspace{0.7pt} \beta_{0,t} + n_e - \sum \textbf{z}_t)
\end{equation}
where $beta$ is the Beta distribution and $\alpha_{0,t}$ and $\beta_{0,t}$ define the Beta distribution prior. We here model the prior in terms of the mean ($\mu_t$) and variance ($\sigma^2_t$) of the distribution:
\begin{equation}
\begin{aligned}
    \alpha_{0,t} = \mu_t \left( \frac{\mu_t(1-\mu_t)}{\sigma^2_t}-1 \right) \\
    \beta_{0,t} = (1-\mu_t) \left( \frac{\mu_t(1-\mu_t)}{\sigma^2_t}-1 \right) \\
\end{aligned}
\end{equation}

Our Beta prior is a heuristic that reflects the intuition that the confidence should be high for times with near-zero variability (i.e., no corrections) and uncertain otherwise (i.e., $p_{c\mid t}\approx\frac{1}{2}$):
\begin{equation}
\begin{aligned}
    \mu_t = \frac{1}{2}\Big(1-e^{-\gamma_p \left( \sigma^2_{\mathcal{D}}(t) +\epsilon\right)}\Big) \\
    \sigma^2_t = \sigma^2_{\textrm{MAX}}\Big(1-e^{-\gamma_p \left( \sigma^2_{\mathcal{D}}(t) +\epsilon \right)}\Big)
\end{aligned}
\end{equation}
where $\gamma_p$ and $\sigma^2_\textrm{MAX}$ are scaling parameters for the prior and $\epsilon$ is a small constant to make the prior well posed. For scheduling, we determine whether the robot is confident, $\conf(t) \in \mathbb{B}$ (i.e., the robot is low or high confidence), from the Beta cumulative distribution function (CDF):
\begin{equation}
    \conf(t) = p(p_{c\mid t} < \mu_c \mid \textbf{z}_t , t) > \gamma_c
\end{equation}
\vspace{-.2cm}
where $\mu_c$ is the probability of corrections where it is acceptable for the robot to execute without supervision and $\gamma_c$ is the corresponding confidence interval of the distribution.

\subsubsection{Scheduling Around Confidence} The full scheduling problem consists of determining a task execution (i.e., time warp), $\psi_{\mathcal{E}_i}$, for each robot and a start offset for the second robot, $\tau$. Each execution warp, $\psi_{\mathcal{E}_i}$, is defined relative to the reference demonstration (similar to the demonstrations) and scheduled by choosing permissible values of the warp gradient. We desire to schedule the execution warps and offset to avoid overlapping low-confidence actions and to minimize the expected total time, $T_{\textrm{total}}$, for both robots to finish the task.
\begin{equation}
 T_{\textrm{total}} = \max \{ |\psi_{\mathcal{E}_1}|, \tau + |\psi_{\mathcal{E}_2}| \}
\end{equation}
where $|\psi_{\mathcal{E}_i}|$ denotes the length of the execution warp (i.e., the duration of the behavior). For each robot, we determine a warped execution from the mean warp gradient, $ \dot{\bar{\psi}}(t)$, and the minimum and maximum demonstration warp gradients.
\begin{equation}
    \begin{aligned}
    \dot{\psi}^{\textrm{min}}(t) = \min \{ \dot{\psi}_{\mathcal{D}_1}(t), \ldots, \dot{\psi}_{\mathcal{D}_{n_d}}(t) \} \\
    \dot{\psi}^{\textrm{max}}(t) = \max \{ \dot{\psi}_{\mathcal{D}_1}(t), \ldots,  \dot{\psi}_{\mathcal{D}_{n_d}}(t) \}
    \end{aligned}
\end{equation}
where $\dot{\psi}^{\textrm{min}}(t)$ and $\dot{\psi}^{\textrm{max}}(t)$ are the minimum and maximum warp gradients respectively. 

When the robot is executing a low-confidence action with operator corrections, the warps should follow the mean warp gradient and mean behavior. When the robot is executing a high-confidence action, we have the flexibility to warp the timing within the demonstration bounds to increase the likelihood of finding a non-overlapping scheduling solution. While it may be possible in some tasks to warp the executions beyond what was seen in the demonstrations (e.g., stop certain robots, replan between critical regions), we choose to respect the demonstration bounds to limit violating any latent task constraints (e.g., time required to start a tool before beginning a critical region, maximum time before a sealant dries). 

To this point, we have discussed timing in the context of the reference demonstration. Our formulation ultimately creates new behavior warps and schedules them in a coordinated global fashion. To simplify the notation in our formulation, we introduce a global scheduling notation to convert between times in the overall execution (i.e., both robots) and the corresponding time in a robot's reference behavior.
\begin{equation}
t_{\mathcal{E}_i} (t) =\left\{
\begin{array}{lll}
      0 & t < \boldsymbol{\tau}_i \\
      \psi_{\mathcal{E}_i}^{-1}(t-\boldsymbol{\tau}_i)  & \boldsymbol{\tau}_{i} \leq t \leq  \boldsymbol{\tau}_{i}+|\psi_{\mathcal{E}_i}|\\
       \psi_{\mathcal{E}_i}^{-1}(|\psi_{\mathcal{E}_i}|) & else 
\end{array} 
\right.
\end{equation}
where $\psi_{\mathcal{E}_i}^{-1}$ is the inverse warp that goes from the execution to the reference demonstration and  $\boldsymbol{\tau}$ is a vector of offsets with the first offset set to zero (i.e., $\boldsymbol{\tau} = [ 0,  \tau ]$). With this global reference time, it is possible to assess the confidence of each robot at any time (set to one when not running) or both robots.
\begin{equation}
\begin{aligned}
    \conf_{i}(t)=\left\{
\begin{array}{ll}
      \conf( t_{\mathcal{E}_i} (t)) & \boldsymbol{\tau}_{i} \leq t \leq  \boldsymbol{\tau}_{i}+|\psi_{\mathcal{E}_i}|\\
      1 & else 
\end{array} 
\right. \\
\bconf(t)= [ \conf_{1}(t), \conf_{2}(t) ]
\end{aligned}
\end{equation}

\algnewcommand\algorithmicnot{\textbf{not}}
\algnewcommand\algorithmicand{\textbf{and}}
\algnewcommand{\algorithmicor}{\textbf{ or }}
\algnewcommand{\OR}{\algorithmicor}
\algnewcommand{\AND}{\algorithmicand}
\algnewcommand{\NOT}{\algorithmicnot}
\algdef{SE}[IF]{IfNot}{EndIf}[1]{\algorithmicif\ \algorithmicnot\ #1\ \algorithmicthen}{\algorithmicend\ \algorithmicif}%

\algtext*{EndWhile}
\algtext*{EndIf}

\begin{algorithm}[b]
	\caption{Strategy For Two High-Confidence Robots}
 \label{alg:highconf}
	\begin{algorithmic}[1]
            \State $a \leftarrow$ ahead robot (or either if no deviation)
            \State $b \leftarrow$ behind robot
	    \Function{high-conf}{$t_a,t_b,i_a,i_b,\cdot$}
            \State $\Delta T_{ab} \leftarrow t_a - t_b$
            
            \If{$\dot{\psi}_{i_b}^{F}(t_b)\Delta t_s \ge  \dot{\psi}_{i_a}^{F}(t_a)\Delta t_s + \Delta T_{ab}$} \\ \Comment{behind robot can overtake the ahead robot}
                \State $\delta t_b = \dot{\psi}_{i_a}^{F}(t_a)\Delta t_s + \Delta T_{ab}$
                \State $\delta t_a = \dot{\psi}_{i_a}^{F}(t_a)\Delta t_s$
            \Else \Comment{behind robot cannot overtake ahead robot}
                \State $\delta t_b = \dot{\psi}_{i_b}^{F}(t_b)\Delta t_s$
                \State $\delta t_a = \delta t^{\textrm{resp}}_{i_a}(\Delta T_{ab}-\delta t_b, t_a)$
            \EndIf
            \State \Return $\delta t_a, \delta t_b$
	\EndFunction
	\end{algorithmic} 
\end{algorithm}

\subsubsection{Real-time Adaptation to Operator Corrections}
One of the major challenges with scheduling in the shared autonomy setting is that operator corrections can affect the timing of the robots. Given that we know the bounds of operator corrections and the corresponding impacts on timing, we desire to find a scheduling solution that avoids overlapping low-confidence regions under any permissible operator correction.

The simplest way to ensure robustness would be to show that both robots can instantaneously accommodate any timing changes from the corrections (i.e., if one robot slows down or speeds up, the other robot can also slow down or speed up). However, one robot may be completing an action that cannot accommodate the change in real time (e.g., a low-variability sanding pass where there is no flexibility in the speed). In this case, we must show the robots can \emph{make up the timing change} before the next low-confidence region. 

Our strategy uses the execution gradient to determine how much faster or slower the robot can execute the task to make up timing changes. One important point is that the minimum gradient refers to speeding up a behavior and the maximum gradient refers to slowing down the behavior. For example, if $\dot{\psi}_{\mathcal{E}_i}^{\min} = 0.5$, the behavior could run in half as many samples or equivalently could run twice as fast. Thus, to calculate how much faster or slower a robot can go ($F$ and $S$ respectively), we divide the scheduled gradient by the demonstration limits.
\begin{equation}
\begin{aligned}
    \dot{\psi}_{\mathcal{E}_i}^{F}(t) = \dot{\psi}_{\mathcal{E}_i}(t_{\mathcal{E}_i} (t)) ~/~ \dot{\psi}^{\min}(t_{\mathcal{E}_i} (t)) \\
    \dot{\psi}_{\mathcal{E}_i}^{S}(t) = \dot{\psi}_{\mathcal{E}_i}(t_{\mathcal{E}_i} (t)) ~/~  \dot{\psi}^{\max}(t_{\mathcal{E}_i} (t))
\end{aligned}
\end{equation}
For example, if $\dot{\psi}_{\mathcal{E}_i}=3$ and $\dot{\psi}^{\min}=0.5$ at a given time, the robot was scheduled at one third of the reference demonstration speed and could go six times as fast (i.e., $\dot{\psi}_{\mathcal{E}_i}^{F}= 3\;/\;0.5 = 6$).

For the real-time strategy, we consider each robot's strategy at discrete time steps, $t_n$. We assume that the robots are commanded at a fixed sample rate, $\Delta t_s$. During execution, corrections will cause the robots to deviate from the original schedule. Thus, we separately track and increment the time of each robot by $\delta t_i$ based on the execution rate selected by the robot strategy (i.e., $t_i \leftarrow t_i + \delta t_i$). The goal of the two robots is to minimize the relative change in timing between the two executions (e.g., $\Delta T_1 = t_1 - t_2$). For brevity, we use a dot within functions (i.e., $f(\cdot)$) to indicate that there are additional previously defined quantities used within the equation. The real-time strategy of the robots can be summarized as follows:
\begin{itemize}[leftmargin=*]
    \item If a robot is executing a low-confidence action, its time increment is determined by the operator's correction, $\delta t^{u}_{i}$. The other robot chooses its response, $\delta t_i^{\textrm{resp}}$, based on the current deviation from the other robot and the bounds of how much it can slow down or speed up. If the responding robot is ahead, it simply goes as slow as possible. If the responding robot is behind, it chooses the move within its limits that minimizes the deviation between the two robots. The full response strategy can be summarized as:
    \begin{equation}
 \delta t^{\textrm{resp}}_{i}(\Delta T_i, t_i)= \left\{
\begin{array}{ll}
      \dot{\psi}^{S}_{\mathcal{E}_i}(t_i) \Delta t_s & -\Delta T_i < \dot{\psi}^{S}_{\mathcal{E}_i}(t_i) \Delta t_s \\
     -\Delta T_i & -\Delta T_i \ge \dot{\psi}^{S}_{\mathcal{E}_i}(t_i) \Delta t_s \; \textrm{and} \\
     & -\Delta T_i < \dot{\psi}^{F}_{\mathcal{E}_i}(t_i)\Delta t_s \\
     \dot{\psi}^{F}_{\mathcal{E}_i}(t_i) \Delta t_s & else
\end{array} 
\right. \\
\end{equation}
    \item When both robots are high-confidence, the robots can plan the fastest move possible while reducing any deviation between the two executions, as described in Algorithm \ref{alg:highconf}. The goal of the robot that is running behind is to go as fast as possible without surpassing the robot that is ahead.
    
    \item If the first robot induces a change to timing while the second robot hasn't yet started, the second robot can simply shift its start time (i.e., increment by the same step). Similarly, if one robot induces a change when the other robot has finished its behavior, there is no need for adaptation.
    
\end{itemize}
 



\begin{figure}[t]
\centering
\includegraphics[width=3.4in]{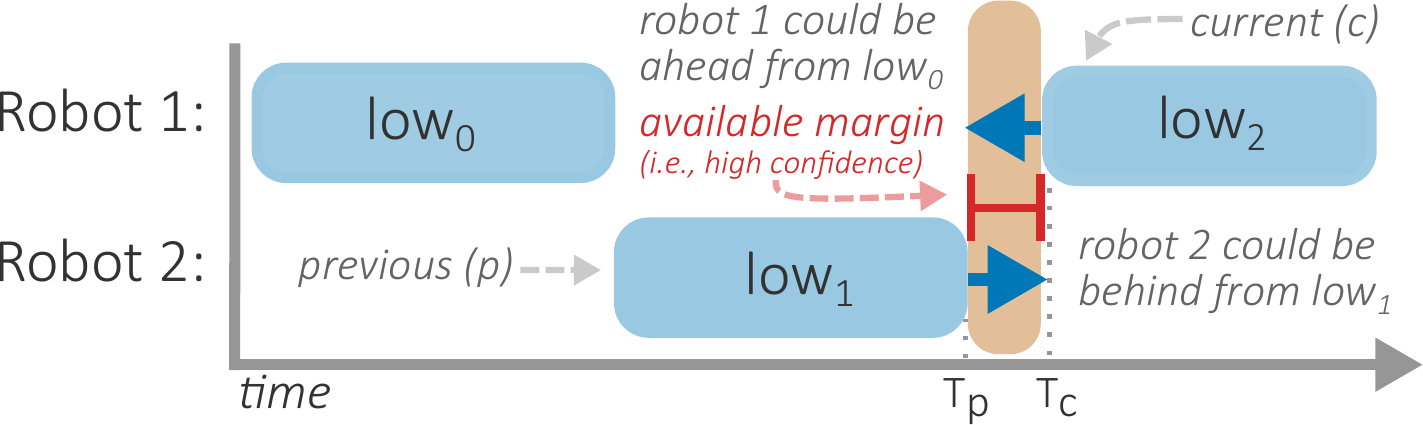}
\caption{Illustrating the required margin (i.e., high-confidence time) between a previous low-confidence region ($low_1$) and current low-confidence region ($low_2$) for a scheduling solution. The worst-case scenario from operator corrections occurs when Robot 1 takes shorter than scheduled on $low_0$ and Robot 2 takes longer than scheduled on $low_1$.}
\label{fig:rtamargin}
\vspace{-15pt}
\end{figure}

With this real-time adaptation strategy, we propose to show that even in the worst case deviation between the two robots, there will still be \textit{no} overlapping low-confidence regions. Consider the example in Figure \ref{fig:rtamargin}. The worst case scenario occurs when the second robot is running behind and the first robot is running ahead (i.e., both behaviors trend toward causing overlap). For each set of adjacent low-confidence regions, we can check if the worst-case behavior leads to the current low-confidence region starting before the previous one ends. The full margin-checking procedure, $\small{\textrm{SUFFICIENT\_MARGINS}}(\cdot)$, is described in Algorithm \ref{alg:margin}. It is important to note that the implementation details, such as a small sample rate in the discretization, impact the validity of the margin calculation.

\subsubsection{Optimization}
Given the required solution properties, we can summarize the multi-robot shared autonomy optimization and solve for the robot execution warps, $\boldsymbol{\psi}_{\mathcal{E}} = \{ \psi_{\mathcal{E}_{1}}, \psi_{\mathcal{E}_{2}} \}$, and offset, $\tau$. The execution warps are chosen based on the permissible warp gradients and defined as an array with the same length as the reference demonstration. To simplify the optimization notation, we define the set of time steps from the reference demonstration and the times in the global notation where a robot is executing a low-confidence action.
\begin{equation}
\begin{aligned}
    \mathcal{T}_{ref} = \{ t_n \in [ 0, |\psi_{D_1}| ] \} \\
    \mathcal{T}_{lc} = \{ t_n \in [ 0, T_{\textrm{total}} ] \mid \min \left( \bconf(t_n) \right) = 0 \}
\end{aligned}
\end{equation}

In our scheduling optimization, the constraints represent strict requirements of the scheduling solution and the minimization quantity expresses additional desiderata for the scheduling solution. The optimization is summarized as:
\vspace{-.1cm}
\begin{align}
    \argmin_{\boldsymbol{\psi}_{\mathcal{E}},\tau} \quad & \textbf{w}^{\intercal}\boldsymbol{\phi}(\cdot) \notag \\
    s.t. \quad & \forall (t_n,\psi) \in \mathcal{T}_{ref} \times \boldsymbol{\psi}_{\mathcal{E}}: \dot{\psi}^{\min}(t_n)\leq \dot{\psi}(t_n) \leq \dot{\psi}^{\max}(t_n) \notag \\
    \quad & 0 \leq \tau \leq |\boldsymbol{\psi}_{\mathcal{E}_1}| \notag \\
        \quad & \forall t_n \in [ 0, T_{\textrm{total}} ]: \max(\bconf(t_n)) = 1  \label{eq:problemformulation} \\
    \quad & \forall (t_n,\psi) \in \mathcal{T}_{lc} \times \boldsymbol{\psi}_{\mathcal{E}}: \dot{\psi}(t_n) = \dot{\bar{\psi}}_{\mathcal{D}}(t_n) \notag \\
    \quad & \small{\textrm{SUFFICIENT\_MARGINS}}(\cdot) = True \notag
\end{align}
where $\boldsymbol{\phi}$ are the costs to minimize and $\textbf{w}$ are the corresponding weights. The constraints enforce that (1) all warps are within the demonstration limits; (2) that the second execution is offset appropriately relative to the first execution; (3) that no unsupervised robot is performing a low-confidence action (i.e., that one of the robots is high-confidence); (4) that the low-confidence robot follows the mean time warping; and (5) that the scheduling solution accommodates variations in timing from corrections. For solutions satisfying the constraints, we can also specify features that differentiate the quality of solutions. In this work, we start with one feature minimizing the total time on task (e.g., $\boldsymbol{\phi}(\cdot) = [ T_{\textrm{total}} ]$). In the future, we will expand the minimization quantity to focus on aspects that improve human factors (e.g., minimizing switching or factoring in switching time for situational awareness) \cite{chen2014human}.

\begin{figure}[t]
\begin{algorithm}[H]
	\caption{Margin Between Low Confidence (LC) Regions}
 \label{alg:margin}
	\begin{algorithmic}[1]
	    \Function {sufficient\_margins}{$\cdot$}
            \State $i_p, T_p, t_{n-1} \leftarrow \varnothing,\varnothing,\varnothing$ \Comment{no previous yet}
	        \For {$t_n \in [0, T_{\textrm{total}} ]$}
                \State $\textrm{start}\_\textrm{LC} \leftarrow !\min(\bconf(t_n)) \And \min(\bconf(t_{n-1}))$
                \State $\textrm{end}\_\textrm{LC} \leftarrow \min(\bconf(t_n)) \And !\min(\bconf(t_{n-1}))$
                \If{$\textrm{start}\_\textrm{LC}$}
                    \State $i_c \leftarrow \argmin(\bconf(t_n))$
                    \State $T_c \leftarrow t_n$

                    \If{$i_p$}
                        \State $ok \leftarrow \Call{check\_margin}{i_p,T_p,i_c,T_c,\cdot}$
                        \If{$!ok$}
                            \State \Return False
                        \EndIf
                    \EndIf
                \EndIf
                \If{$\textrm{end}\_\textrm{LC}$}
                    \State $i_p \leftarrow \argmin(\bconf(t_{n-1}))$
                    \State $T_p \leftarrow t_{n-1}$
                \EndIf
                \State $t_{n-1} \leftarrow t_n$
	        \EndFor
	        \State \Return True
	    \EndFunction
	    \Function{check\_margin}{$i_p,T_p,i_c,T_c, \cdot$}
            \If{$i_p = i_c$} \Comment{same robot, no issue}
                \State \Return True
            \EndIf
      \State $done \leftarrow $False
      \State $t_p, t_c \leftarrow \tau$ \Comment{only times when both robots are running}
      \While{$! done$}
          \If{$\conf_{i_c}(t_c)=0 \OR \conf_{i_p}(t_p)=0$}
          \If{$!\conf_{i_{c}}(t_c)$} \Comment{worst case is faster}
            \State $t_c \leftarrow t_c + \dot{\psi}_{i_c}^{F}(t_c)\Delta t_s$
            \State $t_p \leftarrow t_p + \delta t^{\textrm{resp}}_{i_p}(t_p-t_c, t_p)$
          \Else \Comment{worst case is slower}
            \State $t_p \leftarrow t_p + \dot{\psi}_{i_p}^{S}(t_p)\Delta t_s$
            \State $t_c \leftarrow t_c + \delta t^{\textrm{resp}}_{i_c}(t_c - t_p, t_c)$
          \EndIf
          \Else
          \If{$(t_c-t_p) >0$} \Comment{previous is behind}
            \State $\delta t_c, \delta t_p \leftarrow \Call{high-conf}{t_c,t_p,i_c,i_p,\cdot}$
          \Else \Comment{current is behind or the two are equal}
          \State $\delta t_p, \delta t_c \leftarrow \Call{high-conf}{t_p,t_c,i_p,i_c,\cdot}$
            
          \EndIf
          \State $t_p \leftarrow t_p + \delta t_p$
            \State $t_c \leftarrow t_c + \delta t_c$
          \EndIf
          \If{$t_c \ge T_c$}
            \State \Return False \Comment{possible overlap}
          \EndIf
        \If{$t_p \ge T_p$}
            \State $done \leftarrow \textrm{True}$ \Comment{no overlap}
          \EndIf
          \EndWhile
          \State \Return True
	\EndFunction
	\end{algorithmic} 
\end{algorithm}
\vspace{-30pt}
\end{figure}

\subsection{Implementation of Scheduling}
We solve the scheduling optimization offline prior to each use of the system based on the current confidence estimates of the robot behavior. Given the non-convexity of our scheduling problem and dependencies between our optimization variables, we use the anytime sampling approach described below. During each iteration, we sample the offset based on the length of the first execution and sample the warps from the permissible bounds during high-confidence actions. The warps follow the mean demonstration warp during low-confidence times. The solution that minimizes the total time while meeting the constraints is kept. If no satisfying solution is found, we can fall back the fastest serial execution or the previous scheduling solution. As high-confidence time steps cannot transition back to low-confidence time steps (i.e., cannot be given corrections), the previous scheduling solution will always remain valid.

After each iteration of running the two-robot system, we acquire two task demonstrations and two Boolean arrays of whether corrections were given during low-confidence samples. With these new data, we can shift the mean behavior and append our correction observations.
\begin{equation}
\label{eq:dagger}
\begin{aligned}
    \mathcal{D}^{iter+1} \leftarrow \mathcal{D}^{iter}+ \{ \textbf{X}^{iter+1}_1, \textbf{X}^{iter+1}_{2} \} \\
    \hat{\boldsymbol{\mu}}^{iter+1}(t) \leftarrow \textbf{f}(\mathcal{D}^{iter+1}) \\
    \forall t: \textbf{z}^{iter+1}_t \leftarrow \textbf{z}^{iter}_t + \{ z^{iter+1}_{t,1}, z^{iter+1}_{t,2} \} 
\end{aligned}
\end{equation}
The updated correction observations can be used to update the confidence. In practice, the corrections to each robot occur to fractional samples of the reference robot behavior, thus we interpolate when determining the empirical corrections.

\section{Preliminary Evaluation}
\label{sec:experimentalevaluation}

\begin{figure}[]
\centering
\includegraphics[width=3.4in]{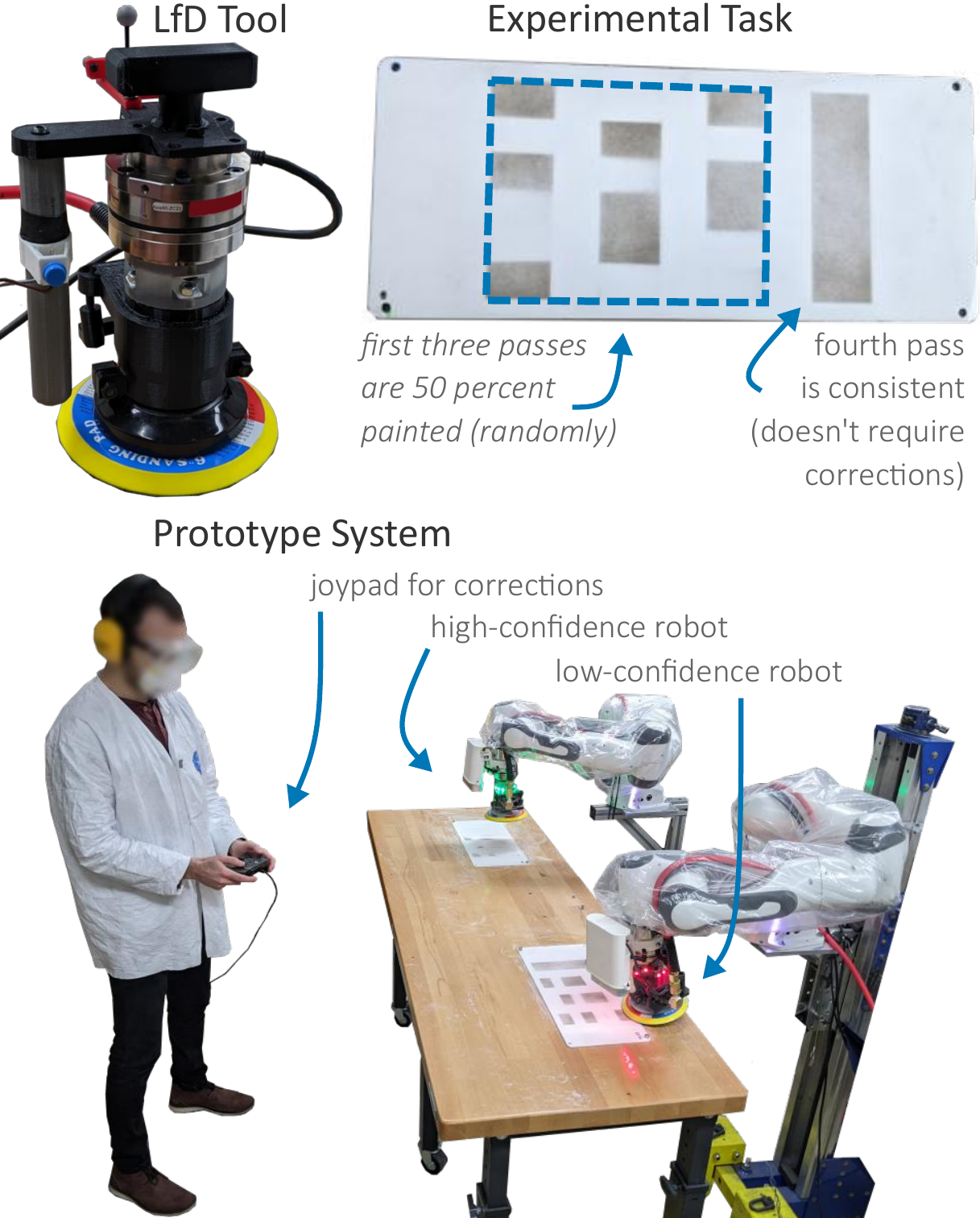}
\caption{Instrumented LfD tool, experiment task, and prototype system.}
\label{fig:expsetup}
\vspace{-15pt}
\end{figure}

As a preliminary assessment of our method, we designed an experimental task, collected demonstrations, and ran a simulated experiment to assess how our scheduling solution adapts with use of the shared autonomy system. The experimental task was inspired by industrial sanding applications (e.g., aircraft and automotive manufacturing) where portions of the task have variable sanding needs. The task consisted of four sanding passes of gray spray paint on a piece of white acrylic, as shown in Figure \ref{fig:expsetup} (top-right). The first three passes were variable, where 50\% of the piece was painted in randomly chosen sections. The fourth pass was consistently painted (i.e., did not require corrections). During demonstrations, the operator applied more force and moved the tool more slowly over the painted sections and applied very little force and moved quickly when there was no paint. As a result, the learned nominal program \textit{lightly} sanded the first three passes and required corrections by the operator to adjust the abrasiveness. During each demonstration, the tool was moved slowly between passes as a simple way to ensure the task was amenable to a highly-parallel execution (i.e., one robot sanding while the other moved between passes). In practice, other high-confidence actions between passes could enable parallel scheduling, such as changing sanding discs. The authors collected a total of six sanding demonstrations using the instrumented sanding tool shown in Figure \ref{fig:expsetup} (top-left). We collected the pose of the tool, the applied force, and the state of the tool (i.e., on/off) during each demonstration.

\begin{figure}[]
\centering
\includegraphics[width=3.4in]{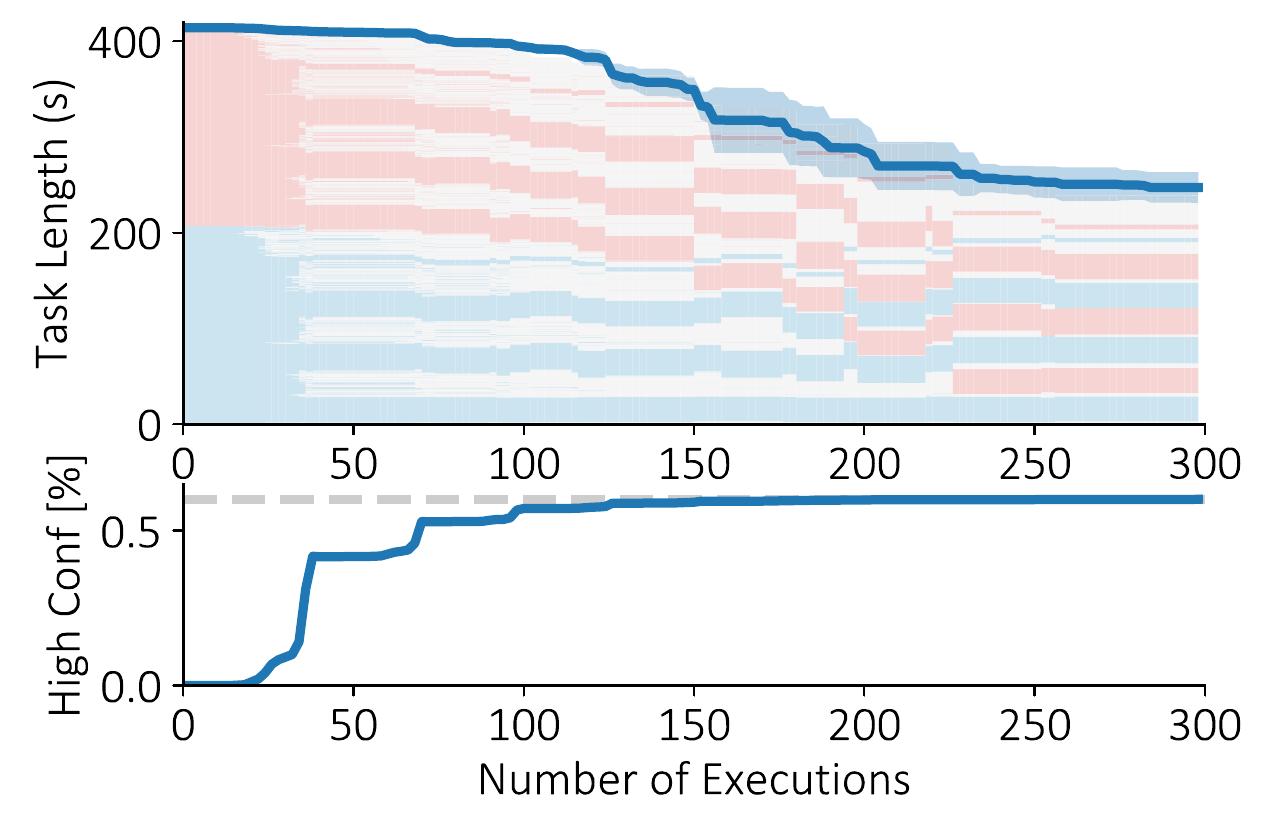}
\vspace{-12pt}
\caption{Scheduling solution evolution. \textit{Top:} The task length decreases over repeated interactions. The dark blue line shows the mean task length (with shading for the standard deviation). For each iteration, we show an example schedule where blue means the first robot is low confidence, red means the second robot is low confidence, and gray means neither robot is low confidence. Initially, the robots are mostly low confidence and execute the task one at a time. Eventually the robots alternate executing low-confidence actions. \textit{Bottom:} Percent of high-confidence samples. The gray dotted line is the ground-truth percentage.}
\label{fig:exp1results}
\vspace{-18pt}
\end{figure}

The simulated experiment consisted of five trials with $300$ simulated executions (in pairs of two) of the system with operator corrections. Consistent with the task, corrections were given $50\%$ of the time on samples during the first three passes. While in the future, we would like to estimate human error rates during corrections to set appropriate confidence bounds, the simulation used a fixed one percent error rate (i.e., the human failed to provide needed corrections $1\%$ of the time and provided unneeded corrections $1\%$ percent of the time). After each pair of executions, the confidence was updated and the schedule was resampled with $5000$ iterations. For our task, the schedule sampling took $3$ minutes ($i5$-$10400F$, $10$ threads). The results, shown in Figure \ref{fig:exp1results}, demonstrate that as the robots receive corrections, the system can refine and leverage its confidence to optimize the task scheduling.

We also developed a prototype system (with an early version of the real-time adaptation), shown in Figure \ref{fig:expsetup}, to conduct a pilot study involving ten participants (3M, 5F, 2 non-binary), aged 18--24 ($M=20.5$, $SD=1.8$), recruited from the UW--Madison campus, under an approved protocol from the university Institutional Review Board (IRB). Participants interacted with the robots serially (one at a time) as a baseline condition and with the final parallel solution found by the scheduler. Primary measures included total task time and paint removal performance, calculated through automated imaging. Results indicate that participants could complete the task with the parallel robots with shorter time on task ($t(9)=25.75,\; p<0.001$) and increased idle time ($t(9)=7.30,\; p<0.001$) without significant impact on paint removal performance ($F(1,9)=2.18,\; p=0.17$). These findings provide a preliminary demonstration of the promise of our approach and a basis for a comprehensive future evaluation.

\section{Discussion}
\label{sec:discussion}
Our preliminary findings suggest that our proposed method can enable users to efficiently share autonomy with multiple robots, though future studies with the prototype system are needed to confirm the benefits. We demonstrated how the scheduling algorithm reduces the time to complete a sanding task as new high-confidence times are identified. As the high-confidence samples approached the ground-truth percentage, our example found a schedule with alternating sanding passes. Generally, our scheduler finds solutions that respect the optimization constraints (e.g., no overlapping, low-confidence actions) and reduces the total time by either speeding up high-confidence actions or overlapping executions when possible.

\subsection{Limitations \& Future Work}
Our method inherits limitations from LfD and time warping, which requires similar demonstrations. To avoid these assumptions, we are interested in exploring a reward-based structure that can better generalize to heterogeneous demonstrations. We also desire to scale our technical methods to larger robot fleets, which requires a modified approach to the margin-checking algorithm presented in Algorithm \ref{alg:margin}, and explore the practical considerations of shared autonomy with larger numbers of agents \cite{chen2014human}. Finally, our preliminary evaluation presented only one example sanding task to highlight the benefits of the proposed method. In the future, we will conduct user studies to assess the empirical human error correction rates and the performance (i.e., accuracy and generalization) of our method across a range of tasks, including different tasks for each robot.



\balance
\bibliographystyle{plainnat}
\bibliography{references}

\begin{thebibliography}{21}
\providecommand{\natexlab}[1]{#1}
\providecommand{\url}[1]{\texttt{#1}}
\expandafter\ifx\csname urlstyle\endcsname\relax
  \providecommand{\doi}[1]{doi: #1}\else
  \providecommand{\doi}{doi: \begingroup \urlstyle{rm}\Url}\fi

\bibitem[Cai et~al.(2022)Cai, Dahiya, Wilde, and Smith]{cai2022scheduling}
Yifan Cai, Abhinav Dahiya, Nils Wilde, and Stephen~L Smith.
\newblock Scheduling operator assistance for shared autonomy in multi-robot
  teams.
\newblock \emph{arXiv preprint arXiv:2209.03458}, 2022.

\bibitem[Chen and Barnes(2014)]{chen2014human}
Jessie~YC Chen and Michael~J Barnes.
\newblock Human--agent teaming for multirobot control: A review of human
  factors issues.
\newblock \emph{IEEE Transactions on Human-Machine Systems}, 44\penalty0
  (1):\penalty0 13--29, 2014.

\bibitem[Crandall et~al.(2010)Crandall, Cummings, Della~Penna, and
  De~Jong]{crandall2010computing}
Jacob~W Crandall, Mary~L Cummings, Mauro Della~Penna, and Paul~MA De~Jong.
\newblock Computing the effects of operator attention allocation in human
  control of multiple robots.
\newblock \emph{IEEE Transactions on Systems, Man, and Cybernetics-Part A:
  Systems and Humans}, 41\penalty0 (3):\penalty0 385--397, 2010.

\bibitem[Dahiya et~al.(2022)Dahiya, Akbarzadeh, Mahajan, and
  Smith]{dahiya2022scalable}
Abhinav Dahiya, Nima Akbarzadeh, Aditya Mahajan, and Stephen~L Smith.
\newblock Scalable operator allocation for multi-robot assistance: A restless
  bandit approach.
\newblock \emph{IEEE Transactions on Control of Network Systems}, 2022.

\bibitem[Deriso and Boyd(2022)]{deriso2022general}
Dave Deriso and Stephen Boyd.
\newblock A general optimization framework for dynamic time warping.
\newblock \emph{Optimization and Engineering}, pages 1--22, 2022.

\bibitem[Gao et~al.(2012)Gao, Cummings, and Bertuccelli]{gao2012teamwork}
Fei Gao, Missy~L Cummings, and Luca~F Bertuccelli.
\newblock Teamwork in controlling multiple robots.
\newblock In \emph{2012 7th ACM/IEEE International Conference on Human-Robot
  Interaction (HRI)}, pages 81--88. IEEE, 2012.

\bibitem[Gombolay et~al.(2013)Gombolay, Wilcox, and Shah]{gombolay2013fast}
Matthew Gombolay, Ronald Wilcox, and Julie Shah.
\newblock Fast scheduling of multi-robot teams with temporospatial constraints.
\newblock 2013.

\bibitem[Gupta and Nadarajah(2004)]{gupta2004handbook}
Arjun~K Gupta and Saralees Nadarajah.
\newblock \emph{Handbook of beta distribution and its applications}.
\newblock CRC press, 2004.

\bibitem[Hagenow et~al.(2021{\natexlab{a}})Hagenow, Senft, Radwin, Gleicher,
  Mutlu, and Zinn]{hagenow2021corrective}
Michael Hagenow, Emmanuel Senft, Robert Radwin, Michael Gleicher, Bilge Mutlu,
  and Michael Zinn.
\newblock Corrective shared autonomy for addressing task variability.
\newblock \emph{IEEE Robotics and Automation Letters}, 6\penalty0 (2),
  2021{\natexlab{a}}.

\bibitem[Hagenow et~al.(2021{\natexlab{b}})Hagenow, Senft, Radwin, Gleicher,
  Mutlu, and Zinn]{hagenow2021informing}
Michael Hagenow, Emmanuel Senft, Robert Radwin, Michael Gleicher, Bilge Mutlu,
  and Michael Zinn.
\newblock Informing real-time corrections in corrective shared autonomy through
  expert demonstrations.
\newblock \emph{IEEE Robotics and Automation Letters}, 6\penalty0 (4):\penalty0
  6442--6449, 2021{\natexlab{b}}.

\bibitem[Hoque et~al.(2022)Hoque, Chen, Sharma, Dharmarajan, Thananjeyan,
  Abbeel, and Goldberg]{hoque2022fleet}
Ryan Hoque, Lawrence~Yunliang Chen, Satvik Sharma, Karthik Dharmarajan, Brijen
  Thananjeyan, Pieter Abbeel, and Ken Goldberg.
\newblock Fleet-dagger: Interactive robot fleet learning with scalable human
  supervision.
\newblock \emph{arXiv preprint arXiv:2206.14349}, 2022.

\bibitem[Ji et~al.(2022)Ji, Dong, and Driggs-Campbell]{ji2022traversing}
Tianchen Ji, Roy Dong, and Katherine Driggs-Campbell.
\newblock Traversing supervisor problem: An approximately optimal approach to
  multi-robot assistance.
\newblock \emph{arXiv preprint arXiv:2205.01768}, 2022.

\bibitem[Losey et~al.(2018)Losey, McDonald, Battaglia, and
  O'Malley]{losey2018review}
Dylan~P Losey, Craig~G McDonald, Edoardo Battaglia, and Marcia~K O'Malley.
\newblock A review of intent detection, arbitration, and communication aspects
  of shared control for physical human--robot interaction.
\newblock \emph{Applied Mechanics Reviews}, 70\penalty0 (1), 2018.

\bibitem[Ravichandar et~al.(2020)Ravichandar, Polydoros, Chernova, and
  Billard]{ravichandar2020recent}
Harish Ravichandar, Athanasios~S Polydoros, Sonia Chernova, and Aude Billard.
\newblock Recent advances in robot learning from demonstration.
\newblock \emph{Annual review of control, robotics, and autonomous systems},
  3:\penalty0 297--330, 2020.

\bibitem[Rosenfeld et~al.(2017)Rosenfeld, Agmon, Maksimov, and
  Kraus]{rosenfeld2017intelligent}
Ariel Rosenfeld, Noa Agmon, Oleg Maksimov, and Sarit Kraus.
\newblock Intelligent agent supporting human--multi-robot team collaboration.
\newblock \emph{Artificial Intelligence}, 252, 2017.

\bibitem[Selvaggio et~al.(2021)Selvaggio, Cognetti, Nikolaidis, Ivaldi, and
  Siciliano]{selvaggio2021autonomy}
Mario Selvaggio, Marco Cognetti, Stefanos Nikolaidis, Serena Ivaldi, and Bruno
  Siciliano.
\newblock Autonomy in physical human-robot interaction: A brief survey.
\newblock \emph{IEEE Robotics and Automation Letters}, 2021.

\bibitem[Swamy et~al.(2020)Swamy, Reddy, Levine, and Dragan]{swamy2020scaled}
Gokul Swamy, Siddharth Reddy, Sergey Levine, and Anca~D Dragan.
\newblock Scaled autonomy: Enabling human operators to control robot fleets.
\newblock In \emph{2020 IEEE International Conference on Robotics and
  Automation (ICRA)}, pages 5942--5948. IEEE, 2020.

\bibitem[Todorov and Jordan(2002)]{todorov2002minimal}
Emanuel Todorov and Michael Jordan.
\newblock A minimal intervention principle for coordinated movement.
\newblock \emph{Advances in neural information processing systems}, 15, 2002.

\bibitem[Yan et~al.(2013)Yan, Jouandeau, and Cherif]{yan2013survey}
Zhi Yan, Nicolas Jouandeau, and Arab~Ali Cherif.
\newblock A survey and analysis of multi-robot coordination.
\newblock \emph{International Journal of Advanced Robotic Systems}, 10\penalty0
  (12):\penalty0 399, 2013.

\bibitem[Zanlongo et~al.(2021)Zanlongo, Dirksmeier, Long, Padir, and
  Bobadilla]{zanlongo2021scheduling}
Sebasti{\'a}n~A Zanlongo, Peter Dirksmeier, Philip Long, Taskin Padir, and
  Leonardo Bobadilla.
\newblock Scheduling and path-planning for operator oversight of multiple
  robots.
\newblock \emph{Robotics}, 10\penalty0 (2):\penalty0 57, 2021.

\bibitem[Zhou and De~la Torre(2012)]{zhou2012generalized}
Feng Zhou and Fernando De~la Torre.
\newblock Generalized time warping for multi-modal alignment of human motion.
\newblock In \emph{2012 IEEE Conference on Computer Vision and Pattern
  Recognition}, pages 1282--1289. IEEE, 2012.

\end{thebibliography}


\end{document}